\documentclass[final]{cvpr}
\usepackage{times}
\usepackage{tabularx}
\usepackage{array}
\usepackage{epsfig}
\usepackage{graphicx}
\usepackage{amsmath}
\usepackage{amssymb}
\usepackage{multirow}
\usepackage{booktabs}
\usepackage{subfigure}  %for multi-graph
\usepackage[table,xcdraw]{xcolor}
\usepackage[ruled]{algorithm2e}
\usepackage{mathtools}
\DeclarePairedDelimiter{\ceil}{\lceil}{\rceil}
\usepackage[pagebackref=true,breaklinks=true,colorlinks,bookmarks=false]{hyperref}

\def\*#1{\mathbf{#1}}
\def\cvprPaperID{6254} % *** Enter the CVPR Paper ID here
\def\confYear{CVPR 2021}

\begin{document}
%%%%%%%%% TITLE
\title{MOOD: Multi-level Out-of-distribution Detection  }

\author{Ziqian Lin$^*$, Sreya Dutta Roy\thanks{Authors contributed equally. Paper is published at CVPR 2021.}, Yixuan Li \\
Department of Computer Sciences\\
University of Wisconsin-Madison\\ 
{\tt \small zlin284, duttaroy, sharonli@wisc.edu}
} 

\maketitle
%%%%%%%%% ABSTRACT
\begin{abstract}
Out-of-distribution (OOD) detection is essential to prevent anomalous inputs from causing a model to fail during deployment. While improved OOD detection methods have emerged, they often rely on the final layer outputs and require a full feedforward pass for any given input. In this paper, we propose a novel framework, multi-level out-of-distribution detection (\textbf{MOOD}), which  exploits intermediate classifier outputs for dynamic and efficient OOD inference. We explore and establish a direct relationship between the OOD data complexity and optimal exit level, and show that \emph{easy} OOD examples can be effectively detected early without propagating to deeper layers. At each exit, the OOD examples can be distinguished through our proposed adjusted energy score, which is both empirically and theoretically suitable for networks with multiple classifiers. We extensively evaluate MOOD across 10 OOD datasets spanning a wide range of complexities. Experiments demonstrate that MOOD achieves up to 71.05\% computational reduction in inference, while maintaining competitive OOD detection performance.
\end{abstract}
%%%%%%%%% BODY TEXT
\section{Introduction}
Out-of-distribution (OOD) detection has become a central building block for safely deploying machine learning models in the real world, where the testing data may be distributionally different from the training data. Existing OOD detection methods commonly rely on a scoring function that derives statistics from the penultimate layer or output layer of the neural network~\cite{HendrycksG17,liang2018enhancing, mohseni2020self,liu2020energy,sehwag2019analyzing,hsu2020generalized}. As a result,
existing solutions require a full feedforward pass for any given test-time input and utilize a fixed amount of computation. This can be undesirable for safety-critical applications such as self-driving cars, where higher computational cost directly translates into higher latency for the model to take prevention in the presence of OOD driving scenes.
Further, the computational cost of OOD detection can be exacerbated by the over-parameterization of neural networks, which nowadays have reached unprecedented depth and capacity. For example, recent computer vision models~\cite{kolesnikov2019big} can have over 900 million parameters, which unavoidably incurs high computational demand during inference time. This motivates the following unexplored question: \emph{how can we
enable out-of-distribution detection that can adjust and save computations adaptively on-the-fly?}

\begin{figure*}
\begin{center}
\includegraphics[width=1.05\textwidth]{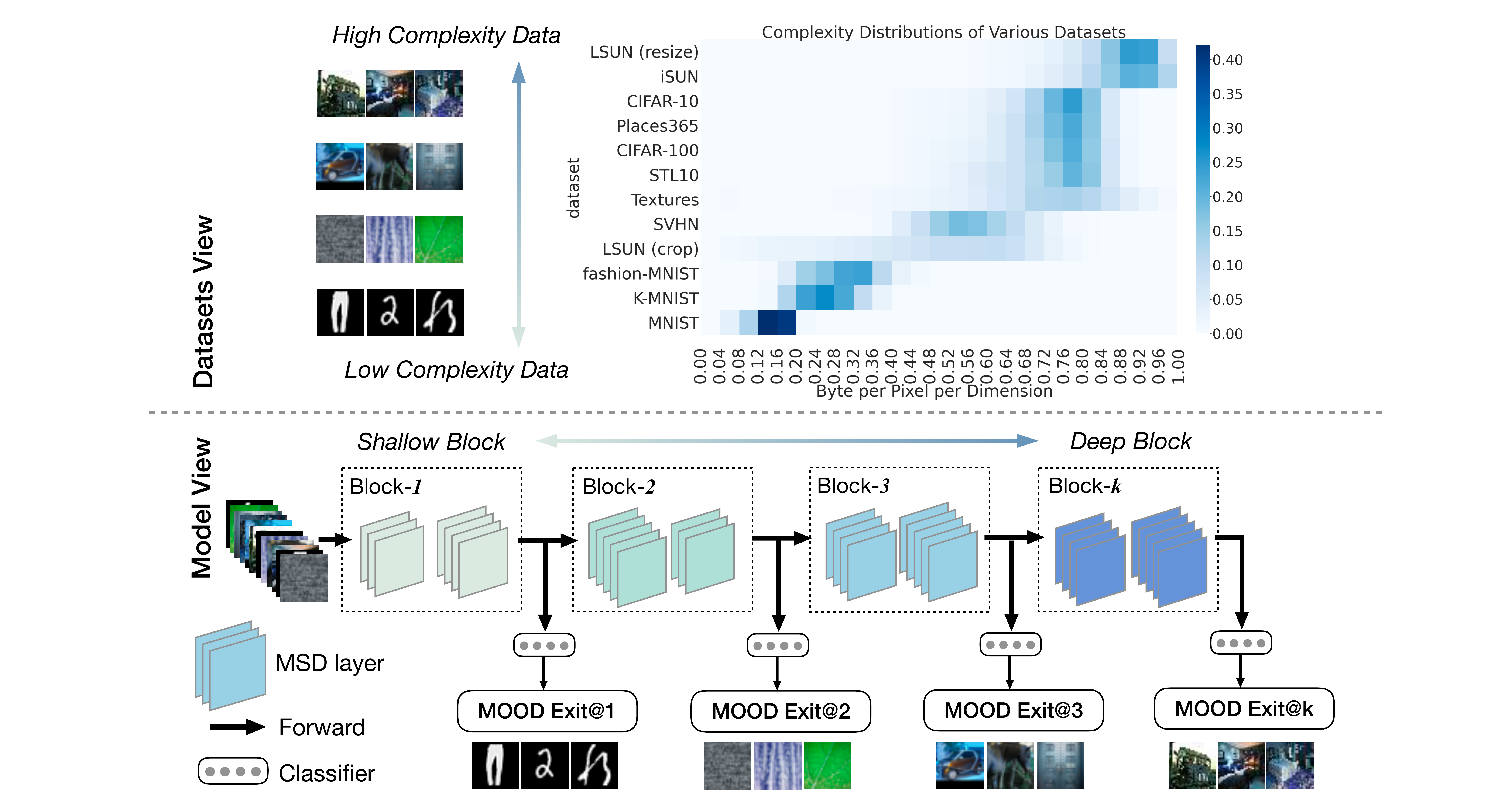}
\caption{\small Overview of proposed \emph{Multi-level Out-of-distribution Detection} (MOOD) framework. MOOD exploits the intrinsic complexity of OOD examples which vary at a wide spectrum (top). The adaptive inference network is composed of $k$ OOD detectors, operating at different depths of the network (bottom). For a given input, a complexity score is used to dynamically determine the exit during inference time. An OOD detector is attached at each exit for differentiating between in- vs. out-of-distribution data.  
\vspace{-0.6cm}
\label{fig:highlevelfig}}
\end{center}
\end{figure*}
In this paper, we take the first step to explore the feasibility and efficacy of an adaptive OOD detection framework based on intermediate classifier outputs. Adaptive OOD detection offers several compelling yet untapped advantages. It allows exploiting the intrinsic complexity of OOD examples which may vary at a wide spectrum (see Figure~\ref{fig:highlevelfig}). Ideally, easy samples could be detected at the early layer, while more complex ones could still be propagated to the deeper layers for more confident decisions. For example, it might be sufficient to utilize coarse-level features such as color to detect an OOD image of grayscale MNIST digit~\cite{lecun2006tutorial}, for a model tasked to classify color images of animals. Secondly, adaptive OOD detection allows flexibility in controlling the computational cost. This is a valuable property in many scenarios, where the computational budget may change over time or vary across different devices.

Formally, we propose a novel framework, {M}ulti-level {O}ut-{o}f-distribution {D}etection (\textbf{MOOD}), which exploits the aforementioned benefits and allows resource-efficient OOD detection. In designing MOOD, we identify three key technical challenges: (1) how to set up an adaptive network with intermediate exits for both OOD detection and classification; (2) how to dynamically choose the optimal exit conditioned on test-time inputs; and (3) at a given exit, how to derive effective OOD scoring function to differentiate between in- vs. out-of-distribution data? 

This paper contributes the following technical components by carefully addressing the three challenges above. 
\begin{itemize}
    \item First, we propose \emph{intermediate OOD detectors} operating at varying depths of the network and enabling dynamic OOD inference. Each intermediate detector is also referred to as an {exit}. Whilst our study is certainly inspired by prior works on resource-efficient learning~\cite{huang2017multi, li2019improved}, the problem we explore differs substantially: most prior solutions are designed for optimizing the classification accuracy for the in-distribution (ID) task, rather than OOD detection. 
    \item Second, we exploit a novel \emph{complexity-based exit strategy}, which uses model-agnostic complexity scores for determining the intrinsic easiness/hardness of the input data. We show a direct relationship between the OOD data complexity and the optimal exit and show that MOOD can detect easy examples early without propagating to deeper layers. Our setting is more challenging than choosing an optimal exit for the image classification task since OOD data are {unexposed} to the model during training.
    \item Thirdly, we introduce an \emph{adjusted energy score} as the OOD scoring function for each intermediate classifier. Our method effectively mitigates the issue of ~\cite{liu2020energy}, where the non-probabilistic energy score can fluctuate and are not comparable across different exits. We show both empirically and mathematically that adjusted energy scores are suitable for MOOD with multiple classifiers operating at different depths of the network. 
\end{itemize}

We extensively evaluate MOOD on a collection of 10 OOD test datasets, spanning a wide range of complexities. We show a relationship between OOD data complexity and the optimal exit level, where the early exit is more favorably chosen for less complex datasets such as \texttt{MNIST} while deeper exits are preferred for complex datasets such as \texttt{iSUN}. Under the same network architecture and capacity, MOOD with dynamic exit reduces the computational cost by up to \textbf{71.05}\%, with an average computation deduction by \textbf{22.02}\% across all 10 datasets. Moreover, MOOD maintains competitive OOD detection performance. Our code and models are available at \url{https://github.com/deeplearning-wisc/MOOD}.
%-------------------------------------------------------------------------
\vspace{-0.3cm}
\section{Method}
\vspace{-0.3cm}
Our proposed adaptive OOD detection framework, \emph{Multi-level Out-of-distribution Detection} (MOOD), addresses three key questions and challenges in designing an adaptive OOD detection framework. Firstly, in Section~\ref{sec:earlyexit}, we describe the adaptive inference model, which allows early exits for OOD detection at varying depths of the network. In Section~\ref{sec:complexity}, we then introduce a novel complexity-based exit strategy, which allows dynamically
determining the optimal exit level for each input during inference time. Lastly, in Section~\ref{sec:energy}, we introduce the OOD inference method operating at different levels of classifiers.

\subsection{OOD Detector at Early Exits}
\label{sec:earlyexit}
We begin by addressing the first challenge of \emph{how do we set up an adaptive inference model with  early exit for OOD detection}?
Inspired by recent works on adaptive neural networks~\cite{bolukbasi17a,huang2017multi,figurnov2017spatially,li2017dynamic,wang2018skipnet,kong2018pixel,mcintosh2018recurrent,wu2018blockdrop,li2019improved},
we consider an adaptive inference model composed of $k$ classifiers. As illustrated in Figure~\ref{fig:highlevelfig}, the model can be viewed as a conventional CNN with $k-1$
intermediate classifiers attached at varying blocks of the network, where each block contains multiple layers. The model can generate a set consisting of $k$ predictions, one
from each of the exits:
\begin{equation}
    f(\*x;\theta) = [f_1(\*x; \theta_1),...,f_k(\*x; \theta_k)],
\end{equation}
where $\theta$ is the parameterizations of the neural network, and $f_i(\*x;\theta_i)$ represents the output from the classifier at exit $i\in\{1,2,...,k\}$. However, prior adaptive networks are designed for optimizing the in-distribution task such as classification or segmentation, and therefore miss the critical component of OOD detection. 

In our framework, we introduce intermediate OOD detectors that operate at each level of classifier and enable dynamic OOD inference. Our key idea is that ``easy" OOD examples can be captured by early layers, whereas more complex ones should be propagated to the deeper layers for more confident decisions. Specifically, each detector $G_i(\mathbf{x})$ can be viewed as a binary classifier:
\begin{equation*}
    G_i(\mathbf{x};\theta_i) = 
    \begin{cases}
    \text{in}, &\text{if}\ S_i(\mathbf{x};\theta_i) \geq \gamma_i \\
    \text{out}, &\text{if}\ S_i(\mathbf{x};\theta_i) < \gamma_i,
    \end{cases}
\end{equation*}
where $S_i(\mathbf{x};\theta_i)$ is the scoring function defined for classifier at exit $i$. $\gamma_i$ is the threshold at exit $i$, which is chosen so that a high fraction (\eg, 95\%) of in-distribution data is correctly classified by the detector. Our multi-level OOD framework allows the full flexibility of early exits at any level, and more importantly, does not incur additional parameterizations on top of the classification networks.

\subsection{Complexity-based Exit for OOD Detection}
\label{sec:complexity}

\begin{figure*}[t!]
\centering
    %\hspace{2mm}
        \includegraphics[width=0.99\textwidth]{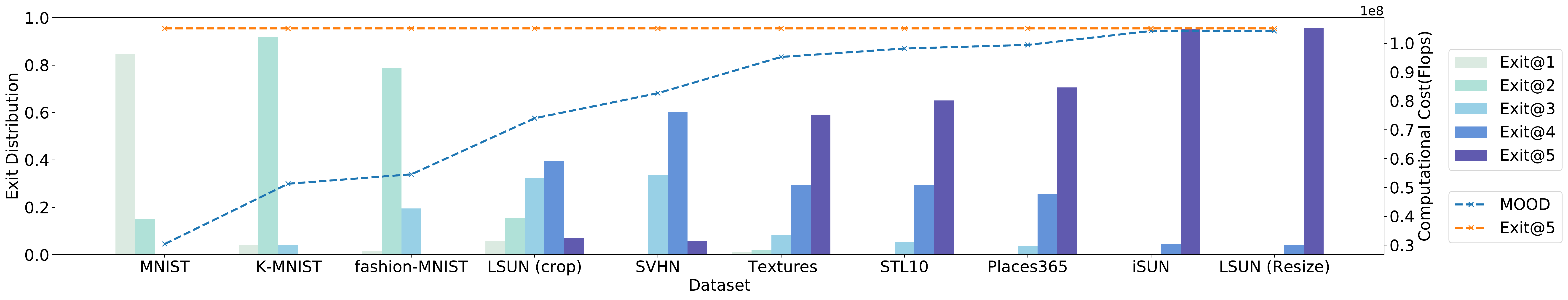}
        \label{fig:flops-barplot}
        \includegraphics[width=0.99\textwidth]{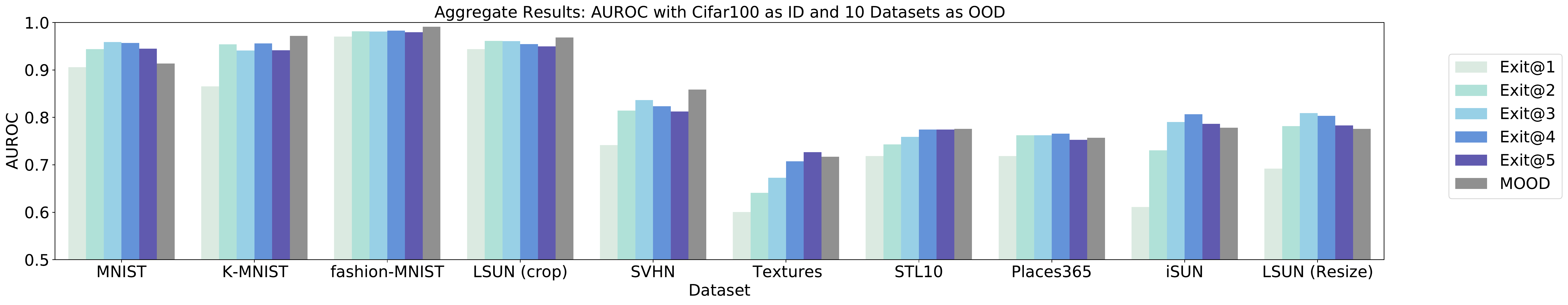}
        \label{fig:auroc-barplot}
    \caption{\emph{Top}: Average computational cost (FLOPs) with MOOD, and the normalized frequency distribution of exits chosen by MOOD. Gap in between the orange and blue lines indicates the computational savings. \emph{Bottom}: Average AUROC by taking constant exits at different levels. Model is trained on CIFAR-100 as in-distribution, and evaluated on 10 OOD test datasets described in Section~\ref{setup}.}
    \label{fig:barplot}
    % \vspace{-0.2cm}
\end{figure*}

In this subsection, we address the second challenge of \emph{how to dynamically choose the optimal exit, conditioned on the test-time inputs?} At test time, we define an exit function  $I(\*x) \in \{1,...,k\}$, which determines the index of the exit classifier for a given input $\*x$. While previous work~\cite{li2019improved, huang2017multi} has relied on prediction confidence to determine the exit for classifying in-distribution data, we argue that such a mechanism is not suitable for OOD detection since the model is not exposed to any OOD data during training. As we show in Section~\ref{fig:barplot}, deeper and more confident layers do not necessarily perform better OOD detection. This suggests the need for an exit strategy that does not depend on the model parameterizations.

To this end, we propose a complexity-based exit strategy and derive an exit function $I(\*x)$ by exploiting the intrinsic complexity of images that naturally varies at a wide spectrum, from easy to hard. In an attempt to quantify this notion of ``easiness", we consider a \emph{complexity score} as a proxy measurement~\cite{serra2019input}. Specifically, the complexity of a given image $\*x$ can be upper bounded by a lossless compression algorithm. An input of high complexity will require more bits while a less complex one will be compressed with fewer bits. The complexity for $\*x$ can be defined as the number of bits used to encode the compressed image:
\begin{align}
L(\*x) = \text{Bit length} (c(\*x)).
\end{align}
For the compression function $c$, one can use common methods such as PNG~\cite{roelofs1999png} and JPEG2000~\cite{skodras2001jpeg}. We further normalize the score by the maximum complexity score of ID dataset, where $L_\text{normalized}=L(\*x)/L_\text{max}$. In Figure \ref{fig:highlevelfig}, we show the distribution of complexity scores for different datasets, ranging from the lowest (\eg, MNIST~\cite{lecun-mnisthandwrittendigit-2010}) to the highest (\eg, LSUN~\cite{yu2015lsun}). 

To make use of the complexity score for inference, we divide the spectrum of complexity scores into $k$ sub-ranges. For any given example $\*x$ (ID and OOD), we assign exit $I(\*x)$ based on the complexity range a sample belongs to:
$$
I(\*x)= \min(\ceil{L_\text{normalized}(\*x) * k}, k).
$$

\begin{figure}[b]
\vspace{-0.3cm}
    \subfigure[Energy Score~\cite{liu2020energy}]{
        \includegraphics[width=0.225\textwidth]{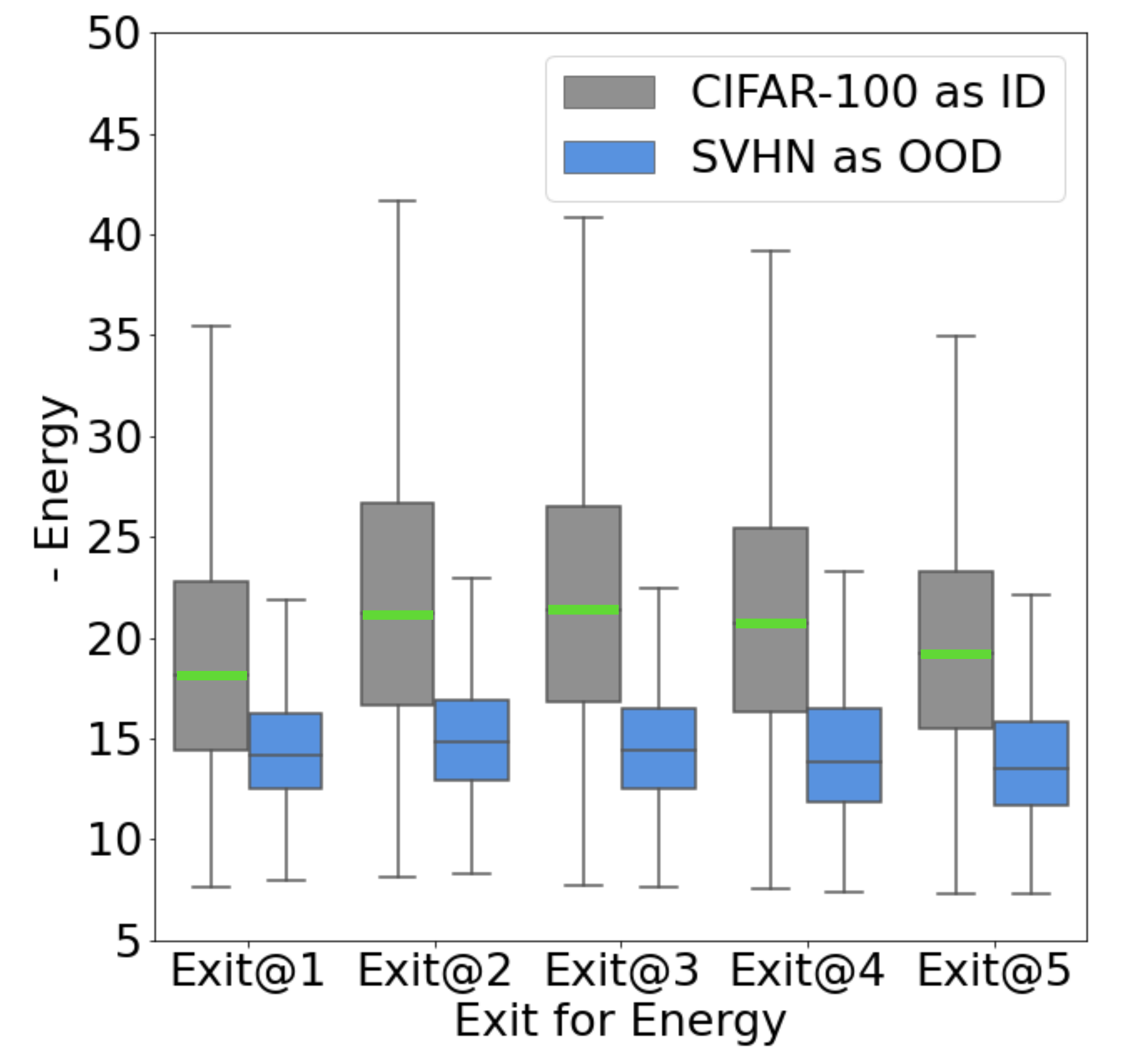}
        \label{fig:RMSE-Lhis}}
    % \subfigure[Adjusted Energy Score.]{
    %     \includegraphics[width=0.227\textwidth]{fig/energy adjusted.pdf}
    %     \label{fig:RMSE-Convs}}
    \subfigure[Adjusted Energy Score (ours)]{
        \includegraphics[width=0.224\textwidth]{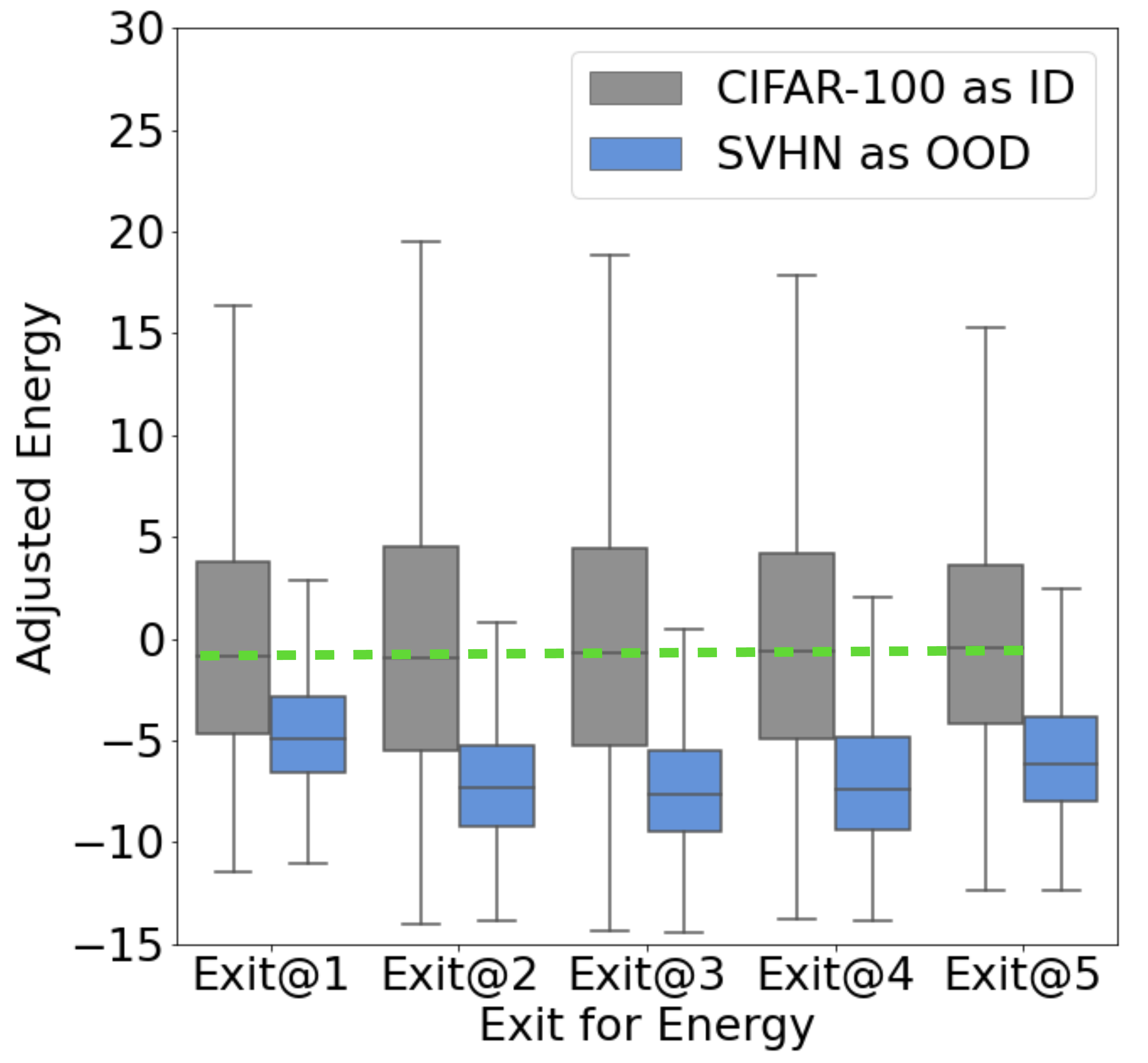}
        \label{fig:RMSE-Convs}}
    \caption{\emph{Left}: Energy scores are not comparable across exits. \emph{Right}: Adjusted energy scores are more comparable across exits (shown in dashed {\color{green}{green}} line).}
    \label{fig:energy_dist}
\end{figure}

\vspace{-0.2cm}
\subsection{OOD Scoring Function}
\label{sec:energy}
 OOD detection is a binary classification problem, which commonly relies on a scoring function to distinguish between in- vs. out-of-distribution data. In this subsection, we address the last key challenge in designing MOOD: \emph{what is the suitable scoring function that allows deriving statistics at intermediate classifiers for OOD detection?}

%\begin{figure}
%\begin{center}
%\includegraphics[width=85mm]{fig/boxdist.png}
%\caption{\small Per Layer Classifier Output Energy Scores for In-distribution and %Out-Distribution samples
%\label{fig:perlayer_classifier_dist}}
%\end{center}
%\end{figure}
%Out-of distribution detection is a binary classification problem and uses a scoring function to help distinguish in-distribution and out-distribution data. 

We begin by exploring the energy-based method for OOD detection, inspired by recent work~\cite{liu2020energy}. Specifically, we derive the \textbf{exit-wise energy score} based on classifier outputs at different exits:
\begin{align}\label{eq:class_energy_softmax}
  E(\*x;\theta_i)= - \log\sum_{j=1}^C e^{f_i^{(j)}(\*x;\theta_i)},
  \vspace{-0.3cm}
\end{align}
where $C$ is the number of classes for in-distribution data, and $f_i^{(j)}(\*x;\theta_i)$ is the logit output corresponding to class $j\in\{1,2,...,C\}$ for intermediate classifier at exit $i$. The likelihood function can be expressed in terms of energy function~\cite{lecun2006tutorial}:
\begin{equation}\label{eq:discriminative}
    p(\*x | \theta_i) = \frac{e^{-E(\*x;\theta_i)}}{\int_{\*x}{e^{-E(\*x;\theta_i)}}}.
\end{equation}
Taking the logarithm on both sides, we have 
\begin{align}\label{eq:logpx}
    \log{p(\*x | \theta_i)} 
    & = -E(\*x;\theta_i) - \log{\int_{\*x}{e^{-E(\*x;\theta_i)}}}\\
    & = -E(\*x;\theta_i) - \log Z_i.
\end{align}
For a single classifier at exit $i$,  the energy score $-E(\*x;\theta_i)$ is indicative of the log likelihood $\log{p(\*x | \theta_i)}$, since the second term $\log Z_i$ is a constant for all $\*x$.

However, in our case with multiple classifiers, the second term $\log Z_i$ cannot be ignored as it depends on exit $i$, and can cause variation across classifiers owing to different input features at each exit. To see this, we depict in Figure \ref{fig:RMSE-Lhis} the energy score distributions at each exit. In particular, the energy scores for CIFAR-100 (ID) shift significantly among exits. Ideally, the scores should be comparable regardless of which exit they come from. %The fluctuation of energy values across exits results in the difficulty of deriving AUROC meaningfully, which ideally requires the scores to be comparable regardless of which exit they come from.

\vspace{-0.2cm}
\begin{algorithm}[t]\label{algo}
\SetAlgoLined
\textbf{Input}: $\*x$, neural network $f(\*x;\theta)$, number of exits $k$, threshold $\gamma$ chosen on in-distribution data\;
\textbf{Compute normalized complexity}: $L_{\text{normalized}} = L(\*x) / L_{\max}$ \;
% \textbf{Choose exit classifier}: $I(\*x)= \mathop{\arg\max}_{i\in\{1,2,...,k\}}(i-1\leq L_\text{normalized}*k)$\;
\textbf{Choose exit classifier}: $I(\*x)= \min(\ceil{L_\text{normalized}(\*x) * k}, k)$\;
% \textbf{Calculate adjusted energy score at exit}: $E_\text{adjusted}(\*x;\theta_{I(\*x)})$\;
\eIf{$E_\text{adjusted}(\*x;\theta_{I(\*x)}) \ge \gamma$}{
  $G(\*x)=\text{in}$\;
  $f(\*x) = f_{I(\*x)}(\*x;\theta_{I(\*x)})$\;
  }{
  $G(\*x)=\text{out}$
  }
\caption{ \textsc{MOOD: Multi-level Out-of-distribution Detection} }
%\vspace{-0.1cm}
\end{algorithm}

\begin{table*}[h]
\centering
\resizebox{0.8\textwidth}{!}{
\begin{tabular}{cclrrrr}
\toprule
\textbf{In-distribution} &
  \textbf{Architecture} &
  \textbf{Method} &
  \textbf{FLOPs} &
  \textbf{AUROC} &
  \textbf{FPR95} &
  \textbf{ID Acc} \\
\textbf{(ID)}  && & $\downarrow$ ($\times 10^8$)  &  $\uparrow$~~~~~ & $\downarrow$~~~~ & ~~~~$\uparrow$ (\%) \\\hline
 &
  &
  MSP~\cite{HendrycksG17} &
  13.00 &
  0.8898 &
  0.5681 &
  94.93\\
  &
WideResNet-40-4     &
  ODIN~\cite{liang2018enhancing} &
  13.00 &
  0.9011 &
  0.3531 &
  94.93\\
  &
    &
  Mahalanobis~\cite{lee2018simple} &
  13.00 &
  0.8933 &
  0.3548 &
  94.93\\
 &
   &
  Energy~\cite{liu2020energy} &
  13.00 &
  0.9004 &
  0.3526 &
  94.93\\ \cmidrule{2-7}
 CIFAR-10 &   
  &
 MSP~\cite{HendrycksG17} &
  1.05 &
  0.8972 &
  0.4987 &
  94.09\\ 
 
 &
MSDNet Exit@last   &
  ODIN~\cite{liang2018enhancing} &
  1.05 &
  0.9033 &
  0.2930 &
  94.09\\  
 
 &
   &
   Mahalanobis~\cite{lee2018simple} &
  1.05 &
  0.8284 &
  0.7519 &
  94.09\\  
 
 &
    &
  Energy~\cite{liu2020energy} &
  1.05 &
  0.9048 &
  0.3362 &
  94.09\\  \cmidrule{2-7}
%  &
%   MOOD &
%   Threshold Based MOOD &
%   0.9009 &
%   0.3073 &
%   2822082 / ... \\
%  &
   &  MSDNet (dynamic exit) & 
   MOOD (\emph{ours}) &
  \textbf{0.79} &
  \textbf{0.9126} &
  \textbf{0.2805} &
  94.13 \\ 
  &&&& ($\pm$0.0016) & ($\pm$0.0051)\\\midrule
 &
 &
  MSP~\cite{HendrycksG17} &
  13.00 &
  0.7710 &
  0.7751 &
  76.90\\
   &
  WideResNet-40-4    &
  ODIN~\cite{liang2018enhancing} &
  13.00 &
  0.8466 &
  0.5722 &
  76.90 \\
  &
   &
  Mahalanobis~\cite{lee2018simple} &
  13.00 &
  0.8319 & 
  0.5352 &
  76.90 \\
   &
    &
  Energy~\cite{liu2020energy} &
  13.00 &
  0.8369 &
  0.6271 &
  76.90 \\ \cmidrule{2-7}
CIFAR-100   &
  &
  MSP~\cite{HendrycksG17} &
  1.05  &
  0.7833 &
  0.7671 &
  75.43\\  

 &
MSDNet Exit@last   &
  ODIN~\cite{liang2018enhancing}  &
  1.05  &
  0.8489 &
  0.5745 &
  75.43\\  
 &
   &
  Mahalanobis~\cite{lee2017training} &
  1.05 &
  0.7380 &
  0.7806 &
  75.43 \\  
 &
  &
 Energy~\cite{liu2020energy}  &
  1.05 &
  0.8451 &
  0.5915 &
  75.43 \\  \cmidrule{2-7}
%   MOOD &
%   MSDNet (dynamic exit) &
%   0.8455 &
%   0.5912 &
%   2822082 / 101240857.5 \\
%  &
   & MSDNet (dynamic exit) & 
  MOOD (\emph{ours}) &
  \textbf{0.79} &
  \textbf{0.8497} &
  \textbf{0.5722} &
  75.26\\ 
  &&&& ($\pm$0.0026) & ($\pm$0.0068)\\\bottomrule
\end{tabular}%
}
\caption{OOD detection performance comparison between MOOD and baselines. All results here are averaged across 10 datasets. \textbf{Bold} numbers are superior results. For MOOD, the complexity calculation takes a negligible amount of time. Therefore, the computations are dominated by the neural network inference cost, measured by FLOPs. The mean and variance of AUROC / FPR95 is reported based on 5 runs. See supplementary A for detailed results for each OOD test dataset.}
\label{table:Comparison WideResNet MSD-Net}
\vspace{-0.3cm}
\end{table*}

% \vspace{-0.3cm}
\paragraph{Adjusted Energy Score} To mitigate this issue, we introduce a new scoring function \emph{Adjusted Energy Score}:
% However, while we have the assumption $p(\*x | \theta_i)$ is only depend on $\*x$ (since data distribution is independent on the model), the denominator $\int_{\*x}{e^{-E(\*x;\theta_i)}}$ is probably varied with $i$, and the $-E(\*x;\theta_i)$ is not comparable among all classifiers. To see this, we depict in Figure \ref{fig:RMSE-Lhis} the energy score distributions at each exit. In particular, the energy scores for CIFAR-100 (ID) shift significantly among exits. 
%The shift is due to that early exit is less converged on in-distribution data and produces outputs that are more conservative, leading to smaller negative energy values. 
%The fluctuation of energy values across exits results in the difficulty of deriving AUROC meaningfully, which ideally requires the scores to be comparable regardless of which exit they come from.
% To resolve this, we propose to use  the \emph{adjusted energy score} defined as below:
%\begin{align*}
%    p(\*x | \theta_i) 
%    &=  \frac{e^{-E(\*x;\theta_i)}}{\int_{\*x}{e^{-E(\*x;\theta_i)}}} \\
%\end{align*}
%define:
\begin{equation}{\label{eqn:defadj}}
     %E_\text{adjusted}(\*x;\theta_i) = \log p(\*x | \theta_i) - \mathbb{E}{[\log p(\*x | \theta_i)]}
     E_\text{adjusted}(\*x;\theta_i) = -E(\*x;\theta_i) - \mathbb{E}_{\*x\in D_\text{in}}{[-E(\*x;\theta_i)]},
\end{equation}
where $\mathbb{E}_{\*x\in D_\text{in}}{[-E(\*x;\theta_i)]}$ is the mean of energy scores derived from exit $i$, and in practice can be estimated empirically on the test set of in-distribution data. We show that \emph{Adjusted Energy Score} produces comparable values across exits. To see this, we can rewrite adjusted energy as:
% \begin{align*}
%   E_\text{adjusted}(\*x;\theta_i) & = (\log e^{-E(\*x;\theta_i)} - \log Z_i) \\ &- (\mathbb{E}_\*x[\log e^{-E(\*x;\theta_i)} - \log Z_i])\\
%   &= \log \frac{e^{-E(\*x;\theta_i)}}{\int_{\*x}{e^{-E(\*x;\theta_i)}}} - \mathbb{E}_\*x[{\log \frac{e^{-E(\*x;\theta_i)}}{\int_{\*x}{e^{-E(\*x;\theta_i)}}}}] \\
%   &= \log p(\*x | \theta_i) - \mathbb{E}_\*x{[\log p(\*x | \theta_i)]}.
% \end{align*}
\newcommand\numberthis{\addtocounter{equation}{1}\tag{\theequation}}
\begin{align*}
   E_\text{adjusted}(\*x;\theta_i) & = (\log e^{-E(\*x;\theta_i)} - \log Z_i) \\& - (\mathbb{E}_\*x[\log e^{-E(\*x;\theta_i)} - \log Z_i]) \numberthis \\
% \end{align*}
% \begin{align*}
    &= \log \frac{e^{-E(\*x;\theta_i)}}{\int_{\*x}{e^{-E(\*x;\theta_i)}}} - \mathbb{E}_\*x[{\log \frac{e^{-E(\*x;\theta_i)}}{\int_{\*x}{e^{-E(\*x;\theta_i)}}}}] \numberthis \\
% \end{align*}
% \begin{align*}
    &= \log p(\*x | \theta_i) - \mathbb{E}_\*x{[\log p(\*x | \theta_i)]}. \numberthis
\end{align*}
% \begin{equation}
%     = -E(\*x;\theta_i) - \mathbb{E}{[-E(\*x;\theta_i)]}\\
% \end{equation}
% \begin{equation}
%     \approx -E(\*x;\theta_i) - \mathbb{E}_{\*x\in D_\text{in}}{[-E(\*x;\theta_i)]}
% \end{equation} \\
\emph{Adjusted Energy Score} is more comparable across exits for all samples $\*x$, since the second term $\mathbb{E}_{\*x\sim D_\text{in}}{[\log p(\*x | \theta_i)]}$ is the estimation of average log-likelihood. The adjusted energy score  can be similar across exits as evidenced in Figure \ref{fig:RMSE-Convs}. The score comparability of in-distribution data allows us to use a \emph{single threshold} $\gamma$ across all exits in inference, which is chosen so that 95\% in-distribution data is above the threshold. %\ZL{In practice, we calculate $\mathbb{E}_{\*x\in D_\text{in}}{[-E(\*x;\theta_i)]}$ based on the entire in-distribution test dataset (\emph{e.g.}, CIFAR-10).}
We summarize the complete inference-time algorithm for MOOD in Algorithm \ref{algo}. For input detected as in-distribution, MOOD also performs classification using the classifier of the current exit. We note that previous methods may require tuning  hyper-parameters~\cite{lee2018simple}, or use auxiliary outlier training data~\cite{HendrycksMD19}. In contrast, MOOD inference is parameter-free and is easy to use and deploy.
\section{Experiments}
We discuss our experimental setup in Section \ref{setup}, and show that MOOD achieves improved OOD detection performance while reducing computational cost by a large margin in Section~\ref{results}. We also conduct extensive ablation analysis to explore different aspects of our algorithm. 

\subsection{Setup} \label{setup}
\paragraph{In-distribution Datasets}
We use CIFAR-10 and CIFAR-100~\cite{krizhevsky2009learning} as in-distribution datasets, which are common benchmarks for OOD detection. We use the standard split, with 50,000 training images and 10,000 test images. All the images are of size $32\times32$. 

\paragraph{Out-of-distribution Datasets}
For the OOD detection evaluation, we consider a total of 10 datasets with a diverse spectrum of image complexity. In order of increasing complexity, we use \texttt{MNIST}~\cite{lecun-mnisthandwrittendigit-2010}, \texttt{K-MNIST}~\cite{clanuwat2018deep}, \texttt{fashion-MNIST}~\cite{xiao2017fashion}, \texttt{LSUN (crop)}~\cite{yu2015lsun}, \texttt{SVHN}~\cite{netzer2011reading}, \texttt{Textures}~\cite{6909856}, \texttt{STL10}~\cite{pmlr-v15-coates11a}, \texttt{Places365}~\cite{zhou2017places}, \texttt{iSUN}~\cite{xu2015turkergaze} and \texttt{LSUN (resize)}~\cite{yu2015lsun}. The complexity distribution of each dataset is shown in Figure~\ref{fig:highlevelfig} (top). All  images are resized to $32\times32$. For each OOD dataset, we evaluate on the entire test split. See supplementary A for details.

\paragraph{Evaluation Metrics}
We evaluate MOOD and baseline methods using the following metrics: (1) Number of computational FLOPS during inference time; (2) False Positive Rate (FPR95) on OOD data when the true positive rate for in-distribution data is $95\%$; and (3) The area under the receiver operating characteristic curve (AUROC).

\paragraph{Training Details}
For training the classification model on in-distribution data, we follow the settings in~\cite{huang2017multi, li2019improved}, and use default MSDNet with $k=5$ blocks with 4 layers each. We use the default growth rate of 6, with scale factors $[1,2,4]$. We train the model for 300 epochs, using  Gradient Equilibrium (GE)  proposed in~\cite{li2019improved}. The initial learning rate is 0.1, which is decayed by a factor of 10 after
150 and 225 epochs. We use SGD (stochastic gradient descent)~\cite{bottou1998online,sutskever2013importance} with  a mini-batch size of 64. We use Nesterov momentum~\cite{nesterov1983method} with  weight 0.9, and a weight decay of $10^{-4}$. In Section \ref{capexp}, we further explore the performance of MOOD with different model capacities by varying the number of layers within each block to be $\{2,3,4,6,8,12\}$.

\vspace{-0.1cm}
\subsection{Results} \label{results}

\subsubsection{How does MOOD compare with existing OOD detection methods?} \label{mainresults}
The evaluation results on CIFAR-10 and CIFAR-100 are summarized in Table~\ref{table:Comparison WideResNet MSD-Net}. All numbers are averaged across 10 OOD datasets described in Section~\ref{setup}. For fair evaluation, we compare with competitive methods in the literature that derive OOD scoring functions from a model trained on in-distribution data and does not require any auxiliary outlier data. In particular, we compare with MSP~\cite{HendrycksG17}, ODIN~\cite{liang2018enhancing}, Mahalanobis~\cite{lee2018simple} as well as Energy~\cite{liu2020energy}. All methods above rely on outputs or features extracted from the final exit and consume the same amount of computations measured by FLOPs. Under the same network  (MSDNet), MOOD with dynamic exit reduces the computational cost (FLOPs) by up to $\textbf{71.05}\%$ on the low-complexity dataset, and reduces the average FLOPs by $\textbf{22.02}\%$ across all 10 datasets. 
In addition to the computational saving, MOOD achieves better OOD detection performance than the baseline methods. This demonstrates that taking early exit is not only computationally %necessary% 
beneficial but also algorithmically desirable for OOD inference. 

We note that WideResNet~\cite{zagoruyko2016wide} used in prior works is much deeper and computationally more expensive (with $8.9$ million parameters), whereas MSDNet with $2.8$ million parameters achieves similar OOD detection performance, as shown in Table \ref{table:Comparison WideResNet MSD-Net}.

\begin{table}[h]
\centering
\small
\vspace{-0.1cm}
                \begin{tabular}{lcrr}
                        \toprule
    \textbf{Method} & \textbf{SVHN}  & \textbf{CIFAR-100}  & \textbf{CelebA}   \\
 \hline
                         Glow~\cite{kingma2018glow}
                         & 0.64 & 0.65 & 0.54\\
                         JEM~\cite{grathwohl2019your}*  & 0.89 & \textbf{0.87} & 0.79\\
                         Serrà et al~\cite{serra2019input}* & 0.95 & 0.74 & 0.86 \\
                         MOOD & \textbf{0.96} & 0.84 & \textbf{0.88} \\

 \bottomrule
                \end{tabular}
        \caption[]{ Comparison with generative-based models for OOD detection. Values are AUROC\@. In-distribution dataset is CIFAR-10, which is common setting used in all baselines. * indicates the variant with best results for the methods.}
        \label{tab:ebm-results}
        \vspace{-0.2cm}
\end{table}

We also compare with generative-based OOD detection approaches, namely Glow~\cite{kingma2018glow}, JEM~\cite{grathwohl2019your}, and likelihood ratio-based method~\cite{serra2019input}. As shown in Table~\ref{tab:ebm-results}, MOOD achieves competitive OOD detection performance with only ${0.79\times10^8}$ FLOPs, as opposed to Glow-based models~\cite{kingma2018glow} (${4.09\times10^{9}}$ FLOPs) or PixelCNN-based model in~\cite{serra2019input} (${2.78\times10^{10}}$ FLOPs).

\subsubsection{What is the effect of complexity-based dynamic exiting strategy?}
To better understand MOOD's exiting behavior, we show in Figure~\ref{fig:barplot} (top) the normalized frequency distribution of exits chosen by MOOD, for each OOD dataset. We show that early exits are more frequently chosen for less complex datasets such as \texttt{MNIST}, while deeper exits are chosen more often for complex datasets such as \texttt{iSUN}. The blue curve shows the average FLOPs consumed by MOOD for each dataset. This trend signifies a direct relationship between computational cost (FLOPs) and OOD data complexity, which is effectively exploited by MOOD. 

Figure \ref{fig:barplot} (bottom)  depicts the average AUROC by taking constant exits at $\{1,2,...,5\}$ for each OOD dataset. The best AUROC chosen among 5 exits can be viewed as the performance upper bound of MOOD, which chooses exit dynamically based on input complexity. We show that early exit is optimal for less complex datasets such as \texttt{fashion-MNIST} while deeper exits are preferred by more complex datasets such as \texttt{iSUN}. Overall, MOOD with dynamic exiting achieves comparable performance as the best possible exit, on all 10 datasets. This trend signifies a direct relationship between the complexity score and an optimal exit. 
We also show the average results on 10 OOD datasets in Table~\ref{tab:constant}. MOOD overall achieves similar OOD detection performance compared to the best constant exit (upper bound performance), while reducing the computational cost in terms of FLOPs and maintaining a high ID classification accuracy. 

We additionally experiment with an alternative lossless compressor JPEG2000~\cite{skodras2001jpeg} for estimating complexity, which results in similar OOD detection performance (see supplementary B for details).

\begin{table}[t]
\centering
\resizebox{0.48\textwidth}{!}{
\begin{tabular}{lrrrr}
\toprule
& \textbf{FLOPs} & \textbf{AUROC}  & \textbf{FPR95}  & \textbf{ID Acc}\\ 
& ~~~$\downarrow$ ($\times 10^8$) & $\uparrow$ & $\downarrow$ & $\uparrow (\%)$\\
\hline
Exit@1  & {0.267} & 0.7769 & 0.7083 & 65.37 \\
Exit@2  & 0.516 & 0.8313 & 0.5719 & 71.24\\
Exit@3  & 0.689 & 0.8471 & 0.5776 & 74.26\\
Exit@4  & 0.884 & {0.8531} & 0.5712 & 75.10\\
Exit@5  & 1.051 & 0.8451 & 0.5915 & {75.43}\\
\midrule
MOOD (ours) & 0.794 & 0.8507 & {0.5709} & 75.26\\
\bottomrule
\end{tabular}}
\caption{Performance comparison between MOOD (dynamic exit) vs. taking constant exits at different levels (non-dynamic exit). Numbers are averaged over 10 OOD datasets with CIFAR-100 as ID dataset.}
\label{tab:constant}
\end{table}

\subsubsection{Is MOOD compatible with other scoring functions?}
We show that MOOD is a flexible framework that is compatible with alternative scoring functions $S(\*x;\theta)$. To see this, we replace the adjusted energy score with MSP~\cite{HendrycksG17} and ODIN score~\cite{liang2018enhancing} 
%and Mahalanobis distance~\cite{lee2018simple} 
derived from outputs of intermediate classifiers, and report performance in Table~\ref{scoringfunc}. The dynamic exit strategy remains the same, where we utilize the complexity score for determining the exit $I(\*x)$. On CIFAR-10, using MOOD with adjusted energy score results in the optimal performance with average AUROC 91.29\%. While the performance using the ODIN score on CIFAR-100 is slightly better, we note that it requires tuning hyper-parameters. In contrast, MOOD with adjusted energy scores is completely parameter-free and is easy to use and deploy. 

Further, by contrasting the performance with Table~\ref{table:Comparison WideResNet MSD-Net}, we observe that MOOD consistently improves the performance compared to using final outputs (\ie, MSDNet Exit@last), regardless of the scoring function.  

\begin{table}[t]
\centering
\resizebox{0.5\textwidth}{!}{
\begin{tabular}{clll}
\toprule
\textbf{In-distribution} & \textbf{MOOD w/ score} & \textbf{AUROC} ( $\uparrow$)  & \textbf{FPR95} ($\downarrow$) \\ \hline
                & MSP              & 0.8984 & 0.5001 \\
CIFAR-10        & ODIN             & 0.9065 & 0.2906 \\
                & Energy           & 0.9070 & 0.3009 \\
                & Adjusted Energy  & 0.9129 & 0.2834 \\ \midrule
                & MSP              & 0.7974 & 0.7336 \\
CIFAR-100       & ODIN             & 0.8548 & 0.5688 \\
                & Energy           & 0.8433 & 0.6041 \\
                & Adjusted Energy  & 0.8507 & 0.5709 \\ \bottomrule
\end{tabular}}
\caption{Effect of scoring functions in MOOD. All numbers aggregated over 10 datasets.}
\label{scoringfunc}
\vspace{-0.2cm}
\end{table}

\subsubsection{How does model capacity affect MOOD performance?} \label{capexp}
\begin{figure}[h]
    \subfigure[varying layers per block ($k=5$)]{
        \includegraphics[width=0.226\textwidth]{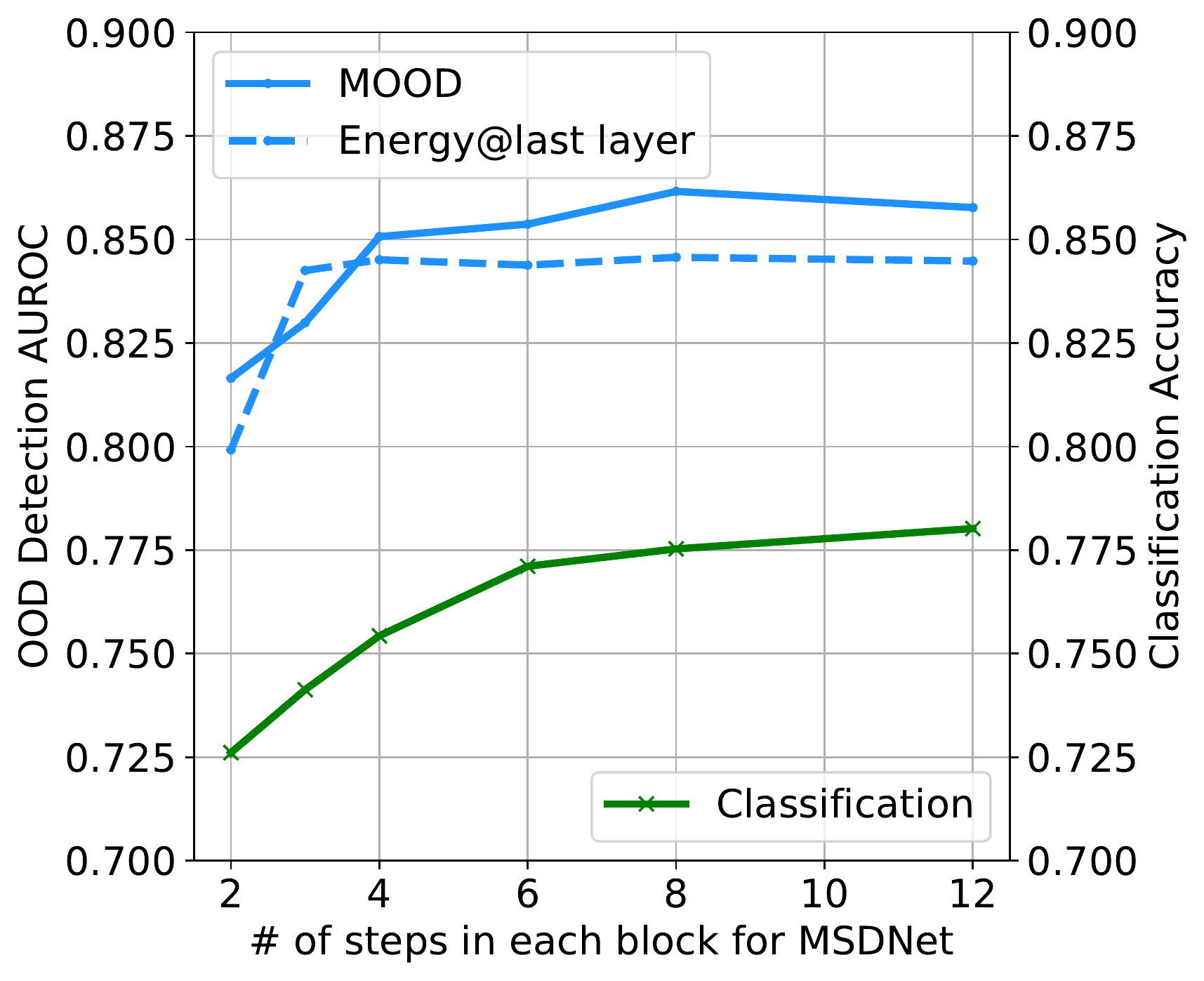}
        \label{fig:layers_cifar100}}
    \subfigure[varying number of blocks $k$]{
        \includegraphics[width=0.2215\textwidth]{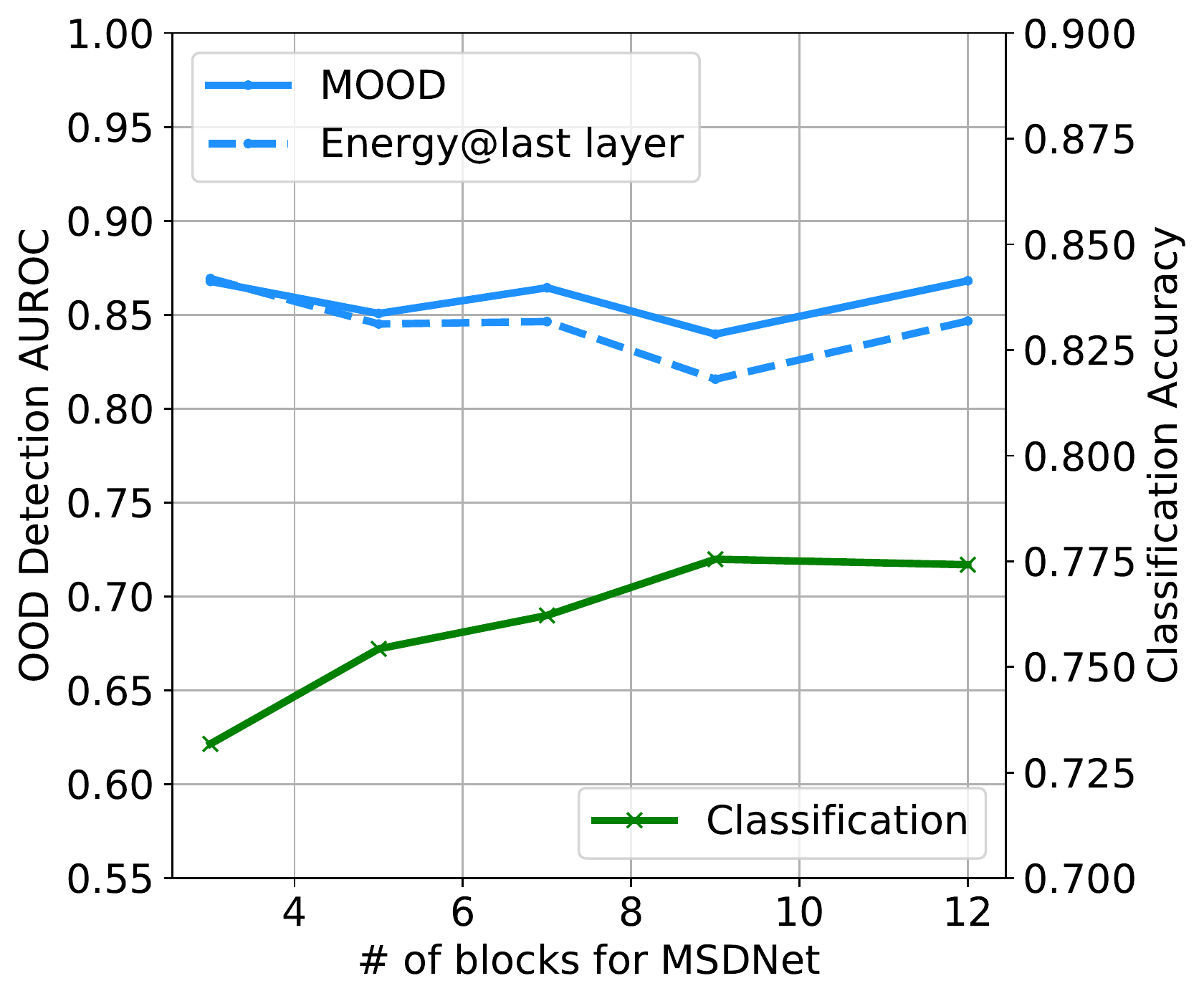}
        \label{fig:block_cifar100}}
    \caption{Effect of model capacity on MOOD performance (CIFAR-100 as ID). \emph{Left}: we fix the number of blocks $k=5$ and vary the number of steps in each block. \emph{Right}: we vary the number of blocks. Blue curves indicate OOD detection performance (AUROC). Green curves indicate classification accuracy on in-distribution data.}
    \label{fig:varycapacity}
\end{figure}

We explore the effect of model capacity on MOOD performance by increasing the depth of MSDNet.  First, we fix the number of blocks $k=5$ and vary the number of steps for each block to be $\{2,3,4,6,8,12\}$, resulting in network configurations with varying total depths. As shown in Figure \ref{fig:layers_cifar100}, we observe that MOOD with dynamic exit works better than the best baseline~\cite{liu2020energy} using the last layer output. Additionally, we explore the effect of varying number of blocks $k\in\{3,5,7,9,12\}$, as shown in Figure~\ref{fig:block_cifar100}. In both cases, the improvement is more pronounced as the network gets deeper. This suggests the applicability of MOOD at different model capacities. 

\subsubsection{How does MOOD affect in-distribution classification accuracy?}

We show in Table~\ref{table:Comparison WideResNet MSD-Net} that MOOD enables efficient OOD detection while maintaining the classification accuracy on in-distribution data. During inference time, we apply complexity-based  strategy to determine the exit $I(\*x)$ for any given in-distribution test input $\*x$, and use the classification predictions made by the corresponding classifier $f(\*x;\theta_{I(\*x)})$. This confirms that our MOOD framework provides several desiderata \emph{altogether}, including (1) safety guarantee against OOD examples, (2) accurate and comparable predictions on in-distribution examples, and (3) computational efficiency during inference. 
\subsubsection{Comparison with Greedy-based and Random Exit Strategy} \label{thresholdSection}
As an ablation on the effect of alternative exit strategies, we attempt a greedy-based method that takes the exit $i$ whenever the adjusted energy score is below the threshold $\gamma_i$. The algorithm is summarized in Algorithm~\ref{algo-greedy}. In contrast to our complexity-based method, the greedy-based strategy depends on the model parameterizations and thresholds for each intermediate OOD exit. $\gamma_i$ is chosen for each exit $i$ so that 95\% of the in-distribution data is correctly classified.

In Table \ref{tab:greedy}, we show results comparison between complexity- vs. greedy-based exiting strategies. The greedy-based method is suboptimal for the task of OOD detection. For our experiment on MSDNet with 5 blocks, the greedy method leads to a $9.26\%$ higher FPR95 compared to our complexity-based method. 
Since the model is not trained with OOD data, the thresholds are less indicative of the optimal OOD exit.

Lastly, we also consider a randomized exit strategy, where we randomly select an exit $\{1,2,...,k\}$ for a given sample during test time. This can be viewed as lower bounding the performance. As expected, the results in Table~\ref{tab:greedy} show a substantial performance drop in both OOD detection as well as in-distribution classification accuracy.

\begin{table}[h]
\centering
\small
\begin{tabular}{lrrrr}
\toprule
\textbf{Exit strategy} & \textbf{FLOPs} & \textbf{AUROC}  & \textbf{FPR95}  & \textbf{ID Acc}\\ 
& $\downarrow$  ($\times 10^8$) & $\uparrow$ & $\downarrow$ & $\uparrow$\\
\hline
Complexity & {0.7943} & \textbf{0.8507} & \textbf{0.5709}  & \textbf{0.7526}\\
Greedy & 0.6890 & 0.8400 & 0.6635 & 0.7384 \\
Randomized & \textbf{0.6880} & 0.8298 & 0.6078 & 0.7299 \\
\bottomrule
\end{tabular}
\caption{Comparison between complexity-based, greedy-based and randomized exiting strategies. Numbers are averaged over 10 OOD datasets with CIFAR-100 as ID dataset.}
\label{tab:greedy}
\end{table}

\begin{algorithm}[t]\label{algo-greedy}
\SetAlgoLined
\textbf{Input}: $\*x$, neural network $f(\*x;\theta)$\;
\While{$i\le k$}{
  Calculate adjusted energy score at exit $i$: $E_\text{adjusted}(\*x;\theta_{i})$\;
  \eIf{$E_\text{adjusted}(\*x;\theta_i) \le \gamma_i$}
  {
   $G(\*x) = \text{out}$\;
   }{
   $i$++\;
  }
 }
$G(\*x)=\text{in}$
\caption{ \textsc{MOOD with Greedy-based Exit} }
\end{algorithm}

%-------------------------------------------------------------------------
\section{Related Work}
\paragraph{Adaptive Inference Networks}
The concept of anytime exit has been previously explored by FractalNets~\cite{larsson2016fractalnet}, deeply supervised networks~\cite{lee2015deeply} and MSDNet~\cite{huang2017multi}.

FractalNets~\cite{larsson2016fractalnet} introduced multiple paths that were evaluated during inference in order of increasing computational requirement. Big-Little Net~\cite{chen2018biglittle} used multiple branches with different computational complexities and merged them to use features of different scales. Dynamic inference methods \cite{bolukbasi17a,figurnov2017spatially,graves2016adaptive,figurnov2017spatially,wu2018blockdrop,veit2018convolutional,wang2018skipnet} adapted the inference to each test instance through gating functions or learning policies, which reduced computation by skipping units or even entire layers. MSDNet~\cite{huang2017multi,li2019improved} also allowed classification for early exiting. Kong and Fowlkes~\cite{kong2018pixel} constructed Pixel-wise Attentional Gating units (PAG) for adaptive inference by learning a dynamic computation path for each pixel. McIntosh \etal~\cite{mcintosh2018recurrent} considered the property of RNNs, varied the number of iterations of the RNN to achieve a flexible range of computational budgets during Segmentation. However, prior solutions are designed for optimizing the in-distribution task such as classification. Our work instead focuses on adaptive inference for OOD detection and introduces %novel%
intermediate detectors operating at different depths of the network.

\paragraph{OOD Detection Using Final Outputs} A baseline method for OOD detection was introduced in~\cite{HendrycksG17}, which uses maximum softmax probability (MSP) from a pre-trained network. Several works attempt to improve the OOD uncertainty estimation using the ODIN score~\cite{liang2018enhancing}, deep ensembles~\cite{lakshminarayanan2017simple}, generalized ODIN score~\cite{hsu2020generalized}, the energy score~\cite{liu2020energy}, and confidence score estimated from a special branch in the model~\cite{devries2018learning}. Several loss functions have been proposed to regularize model predictions of the auxiliary outlier data towards uniform distributions~\cite{lee2017training}, a background class for OOD data~\cite{chen2020informative, mohseni2020self}, or higher energies~\cite{liu2020energy}. However, previous methods have mostly relied on the last layer or the penultimate layer of the neural network. Our paper explores OOD detection by dynamically leveraging intermediate exits, which offers compelling advantages both in computation and detection accuracy.

\paragraph{OOD Detection with Intermediate Information} Several works explored OOD detection using feature representations~\cite{lee2018simple} or outputs~\cite{abdelzad2019detecting} at different intermediate layers of the networks, which signify the usefulness of early layers. Particularly, Lee \etal~\cite{lee2018simple} proposed a scoring function that combined the Mahalanobis distances derived from intermediate features of the neural networks. Abdelzad \etal ~\cite{abdelzad2019detecting} explored exiting at an early layer for OOD detection and required auxiliary dataset to train intermediate OOD detectors. However, in all three methods, there is no adaptive decision made while exiting, and hence computational bottleneck remains unsolved. In contrast, our work takes the first step to explore the feasibility and efficacy of an adaptive OOD detection framework based on intermediate classifier outputs. It is also worth noting that many existing approaches have a number of hyper-parameters that need to be tuned, sometimes with the help of outlier or additional data~\cite{abdelzad2019detecting, lee2018simple}. MOOD with the complexity and adjusted energy scores is parameter-free and does not require any training with auxiliary data.

\paragraph{Complexity and OOD Detection}
Sabeti \etal~\cite{sabeti2019data} combined concepts of typicality and minimum description length to perform novelty detection. They considered atypical sequences that can be described (coded) with fewer bits. 
%Nalisnick et al\cite{nalisnick2018deep} realizes the issue of high confidence likelihoods in generative models when faced with OOD data and introduces complexity along with likelihood in the scoring function. 
Serr\`a \etal~\cite{serra2019input} used input complexity to derive OOD score for generative models such as Glow~\cite{kingma2018glow}, which improved likelihood-based methods~\cite{nalisnick2018deep, ren2019likelihood}. Since it is intractable to compute Kolmogorov complexity~\cite{kolmogorov1963tables}, one can use a lossless compression algorithms such as PNG~\cite{roelofs1999png}, JPEG2000~\cite{skodras2001jpeg} and FLIF~\cite{sneyers2016flif} as a proxy. Different from \cite{serra2019input}, we use complexity in a novel context with multi-level exits and establish the relationship between the complexity of OOD data and suitable exit for dynamic inference.

%-------------------------------------------------------------------------
%%%%%%%%% CONCLUSION
\section{Conclusion}

In this work, we propose a novel framework, \emph{Multi-level Out-of-distribution Detection} (MOOD), which dynamically exploits early classifier outputs for OOD inference. The key idea is to detect easy OOD examples earlier in the network while propagating hard OOD examples to deeper layers. We establish a relationship between OOD data complexity and optimal exit level, which is effectively exploited by MOOD. We extensively evaluate MOOD across 10 OOD datasets spanning a wide range of complexities. We show that MOOD achieves better OOD detection performance than previous approaches relying on final outputs, and at the same time reduces the computations by up to 71.05\%. We hope that future research will increase the attention towards a broader view of the computational efficiency aspect of OOD detection.

%-------------------------------------------------------------------------

{\small
\bibliographystyle{ieee_fullname}
\bibliography{cvpr}
}

\newpage
% % \input{tex/title}
\newpage
\appendix

\begin{center}
      {\Large \bf {MOOD: Multi-level Out-of-distribution Detection
      
      (Supplementary Material)} \par}
     
      % additional small space at the end of the author name
      \vskip .5em
      % additional empty line at the end of the title block
      \vspace*{12pt}
\end{center}

\appendix
\section{Additional Information of Datasets}
\label{supp:dataset info}
In Table~\ref{table:datasetinfo} we provide additional information on the in-distribution and out-of-distribution datasets. We use the entire test splits for each of these datasets and provide the respective test set sizes. Each out-of-distribution input is preprocessed by subtracting the mean of in-distribution data and dividing the standard deviation. MNIST, fashion-MNIST and K-MNIST are padded by 2 in spatial dimensions and then extended to 3 channels. For STL10, SVHN, iSUN, Textures and Places365, the smaller sides of the images are resized to 32 and then center-cropped to 32×32. LSUN (crop) is a dataset created from LSUN by randomly cropping to 32×32 and LSUN (resize) is produced by downsampling each LSUN image to the size 32×32. As previously mentioned, these datasets span a range of complexities, and we present the average complexities per dataset. Table~\ref{table:datasetinfo} also shows the negligible average compression latency per sample while the average inference time of a sample is $20$ ms, using Nvidia 1080 Ti GPU card.

\begin{table}[htbp!]
\centering
\resizebox{0.4\textwidth}{!}{
\begin{tabularx}{\linewidth}{>{\arraybackslash\hsize=1.35\hsize}X>{\raggedleft\arraybackslash\hsize=0.7\hsize}X>{\raggedleft\arraybackslash\hsize=0.85\hsize}X>{\centering\arraybackslash\hsize=1.1\hsize}X}
\toprule
\multirow{3}{*}{\begin{tabular}[c]{@{}c@{}} \textbf{Dataset} \end{tabular}} &
\multirow{3}{*}{\begin{tabular}[c]{@{}c@{}} \textbf{\# of}\\ \textbf{Images} \end{tabular}} &
\multirow{3}{*}{\begin{tabular}[c]{@{}c@{}} \textbf{Mean} \\\textbf{Complexity} \\\textbf{(bytes) }\end{tabular}} &
\multirow{3}{*}{\begin{tabular}[c]{@{}c@{}} \textbf{Average} \\\textbf{Compression} \\\textbf{Latency (ms)} \end{tabular}}\\
& & & \\
& & & \\
\midrule
MNIST         & 10,000                           & 456  & 0.426                                     \\
K-MNIST       & 10,000                            & 799    & 0.480                             \\
fashion-MNIST & 10,000                            & 917    & 0.448                                  \\
LSUN (crop)     & 10,000                           & 1,498     & 0.396                               \\
SVHN          &  10,000                            & 1,736         & 0.339                          \\
Textures      & 5,640                             & 2,165         & 0.348                       \\
STL10         &  8,000                             & 2,222      & 0.338                              \\
CIFAR-100     & 10,000                           & 2,247        & 0.352                           \\
Places365     &  328,500                           & 2,255         & 0.348                      \\
CIFAR-10      & 10,000                           & 2,271       & 0.355                              \\
iSUN          & 8,925                             & 2,690      & 0.350                             \\
LSUN (resize) & 10,000                           & 2,695       & 0.346                                \\ \bottomrule
\end{tabularx}}%}
\caption{Additional information on the 12 datasets listed in the order of increasing complexity. The complexity is measured in bytes after PNG compression. The Compression Latency is measured on Intel(R) Core(TM) i7-7820X CPU @ 3.60GHz and the average inference time of each sample on Nvidia GPU 1080 Ti is  $20$ ms.}
\label{table:datasetinfo}%]
\end{table} 

\section{Experiment on JPEG2000}
\label{supp:other distribution}
\begin{figure}[h]
\centering
        \includegraphics[width=0.4\textwidth]{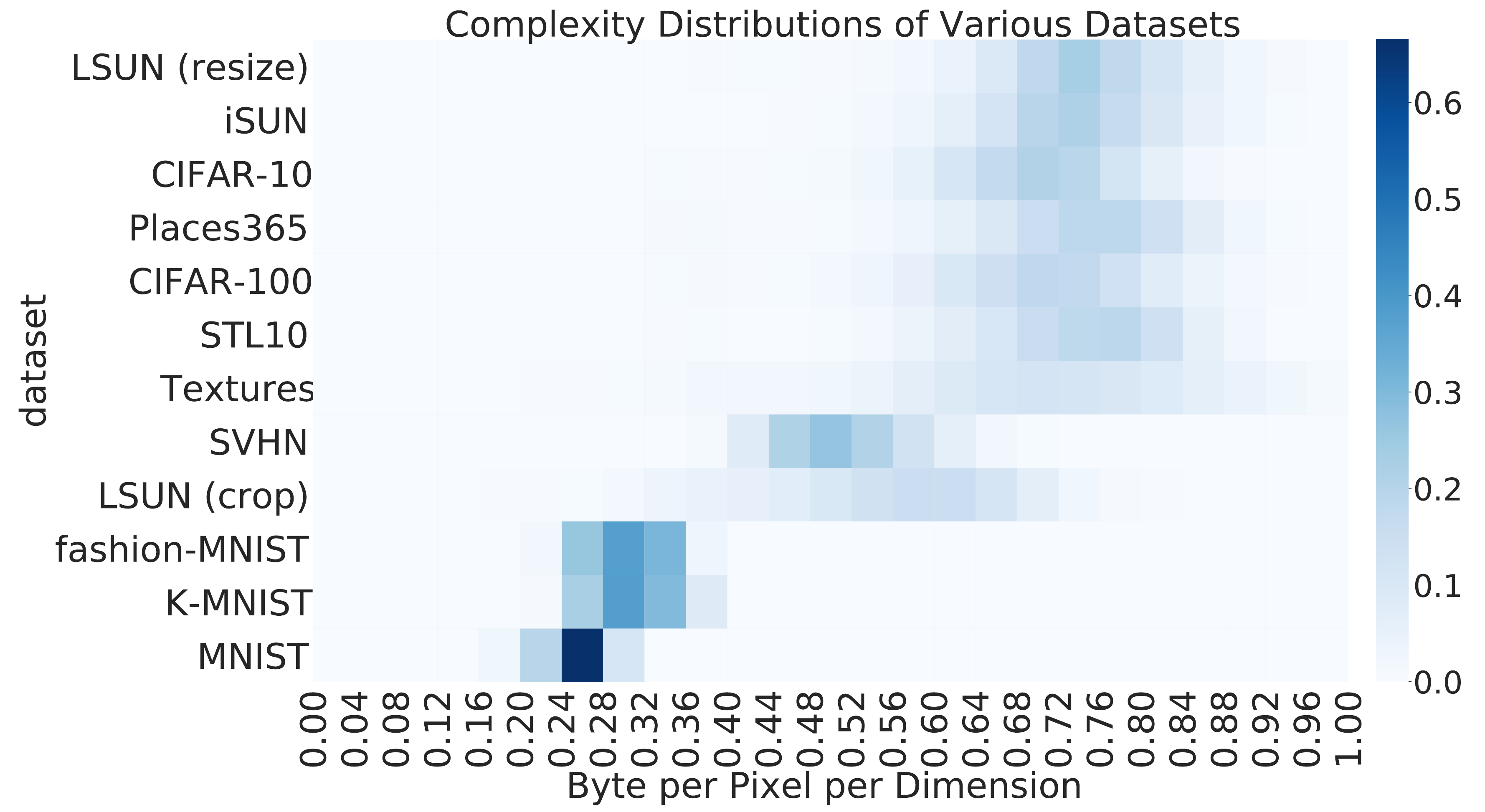}
        \label{fig:heatmap_jpe}
    % \subfigure[CIFAR-100]{
    %     \includegraphics[width=0.8\textwidth]{fig/jpeg.pdf}
    %     \label{fig:#layers_cifar100}}
    \caption{Complexity distribution using JPEG2000. }
    \label{fig:other compressor}
\end{figure}

As seen previously, we choose the PNG compressor for encoding sample images and deriving bit lengths for optimal exit selection in the MOOD algorithm. In this section, we experiment with another lossless compressor JPEG2000. Figure~\ref{fig:other compressor} shows the complexity distribution of samples across the 12 datasets encoded using JPEG2000. 

The result of using JPEG2000 for MOOD is shown in Table~\ref{table:jpeg results}. The JPEG2000 achieves competitive OOD detection results compared with PNG while using more inference time due to the lesser complexity distinguishability of JPEG2000.

\begin{table}[htb!]
\centering
\resizebox{0.4\textwidth}{!}{
\begin{tabular}{cccc}
\toprule
\textbf{ID Dataset}       & \textbf{Method}          & \textbf{AUROC}                & \textbf{FLOPs} \\ 
&&$\uparrow$ & $\downarrow$  ($\times 10^8$)\\
\midrule
\multirow{3}{*}{CIFAR-10}  & Exit@last       & 0.9048               & 1.05  \\
                           & MOOD (PNG)      & 0.9129               & 0.79  \\
                           & MOOD (JPEG2000) & 0.9123               & 0.84  \\ \midrule
\multirow{3}{*}{CIFAR-100} & Exit@last       & 0.8451               & 1.05  \\
                           & MOOD (PNG)      & 0.8507               & 0.79  \\
                           & MOOD (JPEG2000) & 0.8558               & 0.84  \\ \bottomrule
\end{tabular}}
\caption{OOD detection results of JPEG2000 compared to PNG and Exit@last.}
\label{table:jpeg results}
\end{table}
\section{Detailed Results for 10 OOD Datasets}
\label{supp:single dataset}
In Table~\ref{table:all results 1}, we show detailed evaluation results for each of the 10 OOD datasets. We report performance of using both \emph{constant} exiting at each exit, as well as the \emph{dynamic} exit results with our MOOD algorithm.

\begin{table*}[h]
\centering
\resizebox{0.8\textwidth}{!}{
\begin{tabular}{@{}lcccccccc@{}}
\toprule
\textbf{OOD Dataset}              & \textbf{ID Dataset}          & \textbf{Measurement} & \textbf{Exit@1} & \textbf{Exit@2} & \textbf{Exit@3} & \textbf{Exit@4} & \textbf{Exit@5} & \textbf{MOOD} \\ \midrule
\multirow{4}{*}{\textbf{MNIST}}         & \multirow{2}{*}{CIFAR10}  & AUROC                & 0.9744          & 0.9875          & 0.9858          & 0.9889          & 0.9903          & 0.9979        \\
                               &                           & FPR95                & 0.1453          & 0.0546          & 0.0589          & 0.0542          & 0.0413          & 0.0036        \\ \cmidrule(l){2-9} 
                               & \multirow{2}{*}{CIFAR100} & AUROC                & 0.9059          & 0.9440          & 0.9589          & 0.9569          & 0.9451          & 0.9134        \\
                               &                           & FPR95                & 0.5505          & 0.2959          & 0.2491          & 0.2823          & 0.3103          & 0.5770        \\ \midrule
\multirow{4}{*}{\textbf{K-MNIST}}       & \multirow{2}{*}{CIFAR10}  & AUROC                & 0.9800          & 0.9839          & 0.9847          & 0.9868          & 0.9844          & 0.9986        \\
                               &                           & FPR95                & 0.0974          & 0.0805          & 0.0662          & 0.0586          & 0.0699          & 0.0033        \\ \cmidrule(l){2-9} 
                               & \multirow{2}{*}{CIFAR100} & AUROC                & 0.8654          & 0.9539          & 0.9410          & 0.9558          & 0.9416          & 0.9717        \\
                               &                           & FPR95                & 0.7756          & 0.2675          & 0.3616          & 0.2990          & 0.3676          & 0.1663        \\ \midrule
\multirow{4}{*}{\textbf{fashion-MNIST}} & \multirow{2}{*}{CIFAR10}  & AUROC                & 0.9874          & 0.9876          & 0.9912          & 0.9930          & 0.9923          & 0.9991        \\
                               &                           & FPR95                & 0.0548          & 0.0504          & 0.0296          & 0.0219          & 0.0248          & 0.0011        \\ \cmidrule(l){2-9} 
                               & \multirow{2}{*}{CIFAR100} & AUROC                & 0.9705          & 0.9813          & 0.9810          & 0.9827          & 0.9795          & 0.9911        \\
                               &                           & FPR95                & 0.1524          & 0.0843          & 0.1061          & 0.1014          & 0.1226          & 0.0456        \\ \midrule
\multirow{4}{*}{\textbf{LSUN (crop)}}   & \multirow{2}{*}{CIFAR10}  & AUROC                & 0.9796          & 0.9821          & 0.9878          & 0.9877          & 0.9873          & 0.9923        \\
                               &                           & FPR95                & 0.0977          & 0.0953          & 0.0573          & 0.0609          & 0.0591          & 0.0320        \\ \cmidrule(l){2-9} 
                               & \multirow{2}{*}{CIFAR100} & AUROC                & 0.9439          & 0.9613          & 0.9610          & 0.9543          & 0.9495          & 0.9683        \\
                               &                           & FPR95                & 0.2709          & 0.2090          & 0.2176          & 0.2598          & 0.2784          & 0.1702        \\ \midrule
\multirow{4}{*}{\textbf{SVHN}}          & \multirow{2}{*}{CIFAR10}  & AUROC                & 0.8646          & 0.8990          & 0.9391          & 0.9497          & 0.9282          & 0.9649        \\
                               &                           & FPR95                & 0.7600          & 0.5554          & 0.4006          & 0.2892          & 0.3409          & 0.1716        \\ \cmidrule(l){2-9} 
                               & \multirow{2}{*}{CIFAR100} & AUROC                & 0.7418          & 0.8144          & 0.8364          & 0.8238          & 0.8126          & 0.8588        \\
                               &                           & FPR95                & 0.9077          & 0.8120          & 0.7778          & 0.7657          & 0.7756          & 0.6373        \\ \midrule
\multirow{4}{*}{\textbf{Textures}}      & \multirow{2}{*}{CIFAR10}  & AUROC                & 0.8060          & 0.8426          & 0.8732          & 0.8483          & 0.8233          & 0.8332        \\
                               &                           & FPR95                & 0.7259          & 0.6635          & 0.5856          & 0.6016          & 0.5512          & 0.5603        \\ \cmidrule(l){2-9} 
                               & \multirow{2}{*}{CIFAR100} & AUROC                & 0.6003          & 0.6408          & 0.6726          & 0.7073          & 0.7266          & 0.7169        \\
                               &                           & FPR95                & 0.9101          & 0.8780          & 0.8883          & 0.8851          & 0.8690          & 0.8683        \\ \midrule
\multirow{4}{*}{\textbf{STL10}}         & \multirow{2}{*}{CIFAR10}  & AUROC                & 0.6557          & 0.6733          & 0.6757          & 0.6422          & 0.6017          & 0.6131        \\
                               &                           & FPR95                & 0.8479          & 0.8324          & 0.8278          & 0.8438          & 0.8456          & 0.8439        \\ \cmidrule(l){2-9} 
                               & \multirow{2}{*}{CIFAR100} & AUROC                & 0.7185          & 0.7433          & 0.7588          & 0.7743          & 0.7744          & 0.7758        \\
                               &                           & FPR95                & 0.8538          & 0.8273          & 0.8150          & 0.8124          & 0.8131          & 0.7936        \\ \midrule
\multirow{4}{*}{\textbf{Places365}}     & \multirow{2}{*}{CIFAR10}  & AUROC                & 0.8923          & 0.9090          & 0.9128          & 0.8910          & 0.8609          & 0.8674        \\
                               &                           & FPR95                & 0.5004          & 0.4504          & 0.4216          & 0.4547          & 0.4568          & 0.4687        \\ \cmidrule(l){2-9} 
                               & \multirow{2}{*}{CIFAR100} & AUROC                & 0.7187          & 0.7622          & 0.7622          & 0.7656          & 0.7526          & 0.7567        \\
                               &                           & FPR95                & 0.8433          & 0.8014          & 0.8204          & 0.8283          & 0.8265          & 0.8237        \\ \midrule
\multirow{4}{*}{\textbf{iSUN}}          & \multirow{2}{*}{CIFAR10}  & AUROC                & 0.9282          & 0.9612          & 0.9476          & 0.9402          & 0.9384          & 0.9296        \\
                               &                           & FPR95                & 0.3978          & 0.2376          & 0.3190          & 0.3576          & 0.3179          & 0.3882        \\ \cmidrule(l){2-9} 
                               & \multirow{2}{*}{CIFAR100} & AUROC                & 0.6113          & 0.7304          & 0.7901          & 0.8068          & 0.7863          & 0.7784        \\
                               &                           & FPR95                & 0.9248          & 0.8069          & 0.7861          & 0.7394          & 0.7755          & 0.8147        \\ \midrule
\multirow{4}{*}{\textbf{LSUN (resize)}} & \multirow{2}{*}{CIFAR10}  & AUROC                & 0.9409          & 0.9612          & 0.9468          & 0.9450          & 0.9412          & 0.9325        \\
                               &                           & FPR95                & 0.3433          & 0.2400          & 0.3362          & 0.3315          & 0.2911          & 0.3616        \\ \cmidrule(l){2-9} 
                               & \multirow{2}{*}{CIFAR100} & AUROC                & 0.6921          & 0.7816          & 0.8092          & 0.8035          & 0.7832          & 0.7760        \\
                               &                           & FPR95                & 0.8938          & 0.7365          & 0.7542          & 0.7384          & 0.7763          & 0.8122        \\ \midrule
\multirow{4}{*}{Average}       & \multirow{2}{*}{CIFAR10}  & AUROC                & 0.9009          & 0.9187          & 0.9245          & 0.9173          & 0.9048          & 0.9129        \\
                               &                           & FPR95                & 0.3970          & 0.3260          & 0.3103          & 0.3074          & 0.2999          & 0.2834        \\ \cmidrule(l){2-9} 
                               & \multirow{2}{*}{CIFAR100} & AUROC                & 0.7769          & 0.8313          & 0.8471          & 0.8531          & 0.8451          & 0.8507        \\
                               &                           & FPR95                & 0.7083          & 0.5719          & 0.5776          & 0.5712          & 0.5915          & 0.5709        \\ \bottomrule
\end{tabular}}
\caption{Results for 10 OOD datasets. Metrics are AUROC and FPR@95.}
\label{table:all results 1}
\end{table*}

\end{document}